# On the Development of Text Input Method - Lessons Learned

Tian-Jian Jiang[1], Deng Liu[2], Meng-Juei Hsieh[3], and Wen-Lian Hsu[1]

[1] Institute of Information Science, Academia Sinica,
No. 128, Sec. 2, Academia Road,
115 Nan-kang, Taipei, Taiwan
[2] Founder, the OpenVanilla Project
[3] School of Information and Computer Sciences, University of California Irvine

{tmjiang, hsu}@iis.sinica.edu.tw
lukhnos@openvanilla.org
mengjuei@uci.edu

Abstract. Intelligent Input Methods (IM) are essential for making text entries in many East Asian scripts, but their application to other languages has not been fully explored. This paper discusses how such tools can contribute to the development of computer processing of other oriental languages. We propose a design philosophy that regards IM as a text service platform, and treats the study of IM as a cross disciplinary subject from the perspectives of software engineering, human-computer interaction (HCI), and natural language processing (NLP). We discuss these three perspectives and indicate a number of possible future research directions.

## Introduction

To date, most papers on text entry and Input Methods (IM) have focused on automatic conversion from Chinese syllables to words. Chang et al. [1] proposed a system of constraint satisfaction; Kuo [2] developed an application called Hanin using syntactic connection tables and semantic distances; while Hsu [3] presented an application called GOING based on a semantic template matching approach. The latter has become one of the most widely used IMs in Taiwan. In addition, papers indirectly related to automatic Chinese syllable-to-word conversion have been presented at ICCPOL conferences; for example, Tatuoka and LIPS's [4] Vietnamese input system and Zhang's [5] web-based Chinese character input system used in Hong Kong. In this paper, we review these methods and discuss the lessons we have learned from implementing them. Our goal is to share these experiences with both the academic and industrial communities, and thereby help meet the challenges ahead.

The remainder of the paper is divided into three sections: software engineering, human-computer interaction, and natural language processing. We conclude by discussing possible future research directions.

## IM as a Software Engineering Subject

Every modern, GUI-based desktop environment is equipped with sets of API for developing IMs that meet the needs of East Asian markets [6]. However, these API sets are primitive in nature. A developer who wishes to build a fully functional IM system must to handle a myriad of UI events and presentation task. This task has become increasingly difficult for more advanced operating systems, such as Microsoft Windows and Apple's Mac OS X, as the number of features and possible UI events grows. For X11-based desktops, the XIM Framework [7] has long been the standard, but building IM modules with it is no easier than building one for Windows or OS X.

A modern IM framework must have certain features. First, it must provide a set of abstract API. There has to be a dynamic-loading or server-like middle layer between an IM's modules and the underlying operating system, which in turns serves the applications. The framework must also implement a set of platform-specific widgets and event handlers. With such a framework, a third-party IM developer can then concentrate on algorithm design, without being distracted by platform dependent features.

In general, the API of the middle layer should follow the Open-Closed principle [8], which states that software entities (modules) should be open for extensions, but closed to modifications. Being the system software, IM frameworks make extensive use of services provide by modern operating systems. IIIMF is designed with a socket-based client-server architecture, and IM modules must follow its protocol to provide services. OpenVanilla [9] and SCIM [10] are based on the dynamic-loading approach. Their IM modules (as dynamically loaded libraries) and middle-layers stay in the same address space, and modules are loaded. Although such a design reduces the communication complexity substantially, it is limited to a single address space. Thus, modules available on one machine cannot provide services to modules on other machines. Recently, there has been a trend towards client-server based designs. Both OpenVanilla and SCIM now use IPC to talk to their respective "UI servers," though modules are not yet affected.

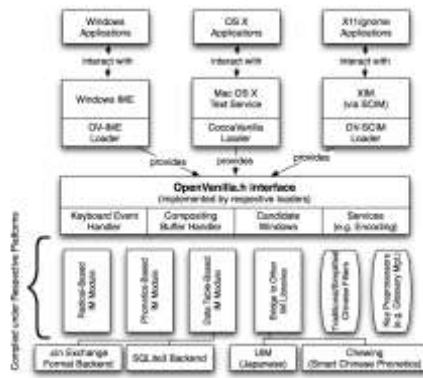

Fig. 1.   Architecture of OpenVanilla platform

Isokoski and Raisamo [11] developed a Java-based client-server architecture that integrates different text entry applications on handheld devices. They compared their work with the Java Input Method Framework (Java IMF) [12], the Microsoft Text Service Framework (MS TSF) [13], and the Internet-Intranet Input Method Framework (IIIMF) [14], and focused on adapting a given text entry application to different devices. However, the authors overlooked an important point about bandwidth: people who want to enter text quickly and continuously find excessive network latency a burden. For someone using Traditional Chinese, a speed of 60 characters per second, with an average of four keystrokes to compose a character, is not uncommon. This means that any response must be within 250 ms, including UI event handling and information retrieval time. Experienced users can type faster; thus, better response times are necessary. Wang and Mankoff [15] proposed an information theory-based model that can quantitatively evaluate relative bandwidth usage. It provides theoretical and architectural support for better device adaptation for IMs with high bandwidth usage. However, despite these developments, the majority of IM modules and frameworks are still limited to a single address space or use the inter-process communication (IPC) schemes at most. The remote procedure call (RPC) based design remains impractical.

Another area of interest is the nature of IMs. East Asian text entry is in fact a series of transformations, whereby several keystrokes are transformed into a character. In software engineering terms, therefore, IM modules can be regarded as filters. However, filtering does not have to stop at keystroke-to-character transformation. We can apply, or even connect, a series of filters so that the output of an IM module can be converted according to the user's needs [9]. One example, which is now a common feature in Chinese IM frameworks, is that users can apply a "Traditional to Simplified Chinese" filter to an IM, so that the script of the typed text is converted on the fly.

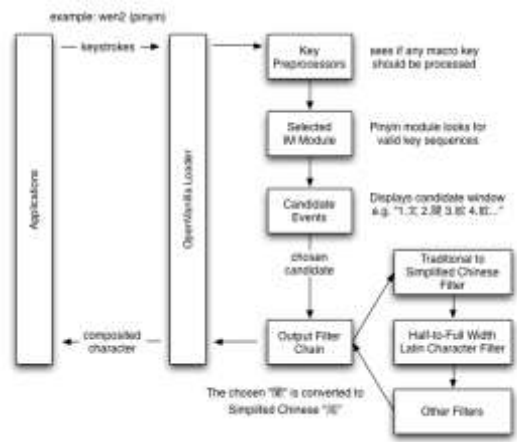

Fig. 2. An IM involves a chain of transformations and events

Even with the IM frameworks cited above, IM modules are very difficult to port and debug. IM frameworks are even more difficult to develop and maintain, as the slowness of porting IIIMF, SCIM, and OpenVanilla demonstrates. Companies such as Microsoft, Apple, and Sun also face the problem of a shortage of engineering expertise in this field, and the complexity posed by a large number of under-maintained legacy codex. This is exemplified by the fact that Apple's Mac OS

X, now a mature system, still keeps an IM component architecture dated back to the mid-1990s [16]. Aside from raising possible security issues, this poses another major development problem in terms of maintenance cost.

## Human-Computer Interaction Principles

In the foreseeable future, the keyboard will remain the primary text entry device, though other concepts such as voice recognition could be potential alternatives. Portable devices, such as mobile phones, present new challenges in keyboard and IM framework design. Fortunately, developments in human-computer interaction (HCI) research can help meet these challenges.

Text entry on handheld devices requires a different approach to keyboard design and layout, since such devices are limited by their size. Mobile phones are the most notable examples. With only twelve numeric keys, even Latin alphabets have to be re-mapped. Fitts' Law [17], which is one of the fundamental principles of HCI research, determines the cost of movements and is usually used to evaluate solutions in this domain. It measures the efficiency of an interface given that users are familiar with basic keyboard functions. One application of Fitts' Law is measuring the best response time based on the keyboard layout and the size of a user's fingers.

A great deal of research in the usability of different text entry methods is based on Fitts' Law, where the variables are keystrokes per character (KSPC), minimum string distance (MSD) or other related metrics [18]. Input methods like T9 [19] or LetterWise [20] are more successful than MultiTap, as they have achieved a balance between the number of keystrokes and the collision rate of keystroke-character conversion. Many users of Traditional Chinese are familiar with Hsu's keyboard layout [ref], which maps Chinese bo-po-mo-fo symbols to 26 keys according to phonetic rules and shape similarity. It is more efficient than any traditional 42-key layout.

For many European users, there are two ways to type characters with diacritic marks. They can either use a language-specific keyboard, or labor with a set of key-combinations that usually involves the use of CTRL or ALT keys. These key combinations are often called the "dead keys," as they are fixed and have to be learned. Given the size limitation of some devices, such as mobile phones, the use of "dead keys" are obviously impractical. During the text entry process, a sequence of keystrokes will probably map to multiple characters, phrases, or "candidates." A user must then pick the exact character/phrase that he or she wants from a list of possible choices. Such interaction requires that candidate characters/phrases be displayed on a screen, called a "candidate list", before a choice can be made. Such a special purpose UI widget is indispensable for Asian language text entry. However, it can have applications other than picking a proper character. In fact, it can serve both as an on-the-fly spelling checker for many European languages and as an alternative to dead keys (an example is to have a choice between "la" and "là" when one types "la"). In other words, a candidate list is a context-sensitive UI widget for any type of text service. Such text-parsing modules have many practical uses. For example, a dictionary/thesaurus agent [9] can offer people writing aides.

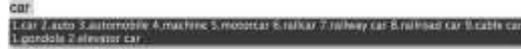

Fig. 3. A WordNet input method module as a thesaurus agent

The design of most text entry methods in handheld devices is based on the same concept of context-sensitive UI widgets. Many designs are now available that use UI widgets to show candidates. A pie menu [21] has proved to be better UI then a linear menu, but only a few PDA devices have implemented it so far. This is due to the difficulty of developing a round widget in most GUI environments and the lack of a pointer device, such as a mouse or a pen. Difficulty also arise when such widgets are deployed on desktop systems as users using keyboards may not like to take an extra step to use a mouse to click on a pie menu. The GOING team has developed a matrix style widget that displays a hierarchical candidate list so that users can choose Chinese words with alphabets rather than numbers (on the forth row of the keyboard). This approach considers both UI widget design and finger movement costs, and only uses half of the alphabets on the left side of an English keyboard. An example of GOING's matrix widget is shown in figure 4.

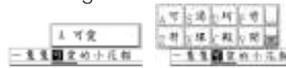

Fig. 4. The second-level candidate windows of GOING

When designing a more accessible interface for elderly or disabled people, simpler movement should be more important than faster typing. The Dasher [22] and the Minimal Device Independent Text Input Method (MDITIM) [23] are two examples of such devices. As mentioned in the previous section, Wang and Mankoff's architecture [15], which provides a model for low bandwidth devices, is also useful for designing interfaces. The model is based on Shannon's noisy channel model [ref], which has been widely adopted in language modeling. In addition, both Dasher and T9 also used simplified uni-gram language models. Clearly, more natural language processing (NLP) techniques are being utilized in HCI research into to design various methods of text entry, which is discussed in the next section.

## How Natural Language Processing Can Enhance Text Entry

In this section we discuss how advances in natural language processing can help the development of better text entry systems, and how such development in turn affects the direction of NLP research.

The semantic template matching approach used in GOING has achieved 95% accuracy. However, the approach is labor-intensive. The GOING team have studied different language modeling techniques and attempted to integrate known semantic templates with GOING in order to take advantage of a number of strategies.

As a syllable-to-word conversion mechanism, an IM can be seen as the last component of automatic speech recognition (ASR) systems that use an n-gram model to predict appropriate words for input keystrokes. For example, Microsoft Pinyin (MSPY; 微軟拼音) [24] and New Zhuyin (新注音) are based on a unified language model in tri-gram and both perform well [25]. Such models, however, are usually quite involved in terms of time and space complexity. Often, a simpler solution can be considered. For example, T9 [19] only adopts a uni-gram model with a smaller set of lexicons. A recent work on Chinese frequent strings [26], showed that it is possible to extract frequent patterns using common information retrieval techniques, such as the Pat-tree algorithm, and then adjust the pattern frequencies to fit the uni-gram model. SCIM's Smart Pinyin (智能拼音) employs a similar approach except that it also implements some heuristic rules of known patterns. Chewing (酷音) [27], which maps bo-po-mo-fo sequences to Chinese characters by matching the longest path in a suffix tree of the directory, can be reprogrammed to use a uni-gram model.

For an IM to adapt to (or learn) the behavior of different users, online learning is necessary. MSPY collects character-based tri-grams to adjust its original model, whereas Chewing updates its suffix tree with an external hash table. The concept behind these adaptation mechanisms is the cache-based language modeling strategy [28]. It is similar to the held-out training method [29]. On the other hand, the adaptive learning [30] approach, which is based on a Bayesian classification, is often used for more sophisticated mechanisms with cached texts. This is also relevant to other areas of research, such as speaker adaptation in the ASR system [31].

In IM research, one of the most important issues is word/phrase identification, which is not easy in any domain or language. A known set of syntactic rules, implemented using LISP, has helped Japanese IM modules detect word boundaries and forms [32]. Since the Chinese language lacks similar deterministic rules, Chinese IMs must depend more on contextual information. To deal with the ambiguity of Chinese word boundaries in language modeling, iterative training procedures [25] have been developed as pure statistical solutions. Following Fred Jelinek's advice to "put language back into language modeling" [33], many researchers have attempted to combine statistical models and linguistic knowledge, especially for long distance linguistic constraints. For example, both the trigger pair [34] and meaningful word pair [35] approaches try to increase the weights of co-occurring grams in language models. Rosenfeld's survey [33], on the other hand, has covered latent semantic analysis, link grammar, dependency grammar and probabilistic context free grammar. These methods are used to build semantic and syntactic knowledge into language models. Models such as maximum entropy [30] and conditional random field [36], currently considered the state-of-the-art, combine all of the above techniques into a unified language model.

It is conceivable that the more complex a model is, the better accuracy can be achieved, likely at the expense of extraordinary computational costs. For most IM implementations, which are expected to be lightweight and highly responsive, these models may not be practical from an engineering point of view. In the next section, we propose several possible solutions to this problem.

## Directions for Future Research and Development

Many Chinese words are single characters, which create a lot of difficulty in the identification of unknown words and unseen events in language modeling. For the latter problem, Chen-Goodman modified Kneser-Ney [29] and other smoothing techniques have been proved useful in Western languages. Yet, they usually fail in oriental languages, because segmented training corpus could often combine these single character words into multiple character words (there is no consistent standard for Chinese word segmentation [1]). The situation can be even worse in the syllable-to-word conversion employed by Chinese IMs, as multiple homonyms can be represented alone or within other lexicons under certain morphological rules. For example, in some phonetic-based "intelligent" IMs, the syllables "yi yang4" are automatically converted into "依樣", whereas "一樣" is expected. This problem is not only affected by the tone features of the Chinese language, but also by word boundaries. In an attempt to resolve this problem, the GOING team used Chen and Goodman's modified Kneser-Ney smoothing technique with word pairs interpolated [35]. With the help of meaningful word pairs, "依樣" is picked only if "畫葫蘆" followed. This is less complicated than the above-mentioned approaches. Preliminary results demonstrate that the syllable-to-word conversion accuracy is improved [35].

Furthermore, the GOING team is experimenting with encoding word pairs and semantic templates into language models based on Bayesian prior probabilities, as suggested by Rosenfeld [33]. The computational cost is expected to be less than that of exponential models or linear discriminant models [37]. To accomplish this, a mathematical model compatible with n-grams and long-distance Bayesian priors must be developed. Our experience in information retrieval research [38] has shown that a hybrid system of Bayesian inference network and language model like Lemur [39] is a good starting point. This concept can be applied to language models for IMs, as well as to online and offline incremental learning mechanisms.

We have integrated various ideas from related research, and applications of network systems. To follow the Open-Closed Principle more closely, an implementation that embeds tiny HTTP daemons (httpd) into the IM platform could be a promising solution. Meanwhile, to avoid the problem in response latency in most thin client architectures, an elegant caching mechanism, similar to the one used in the language model but closer to the hardware architecture design including instruction caching, may be useful. If such a platform were to be implemented, writing IMs in different programming languages other than C/C++ would no longer be difficult, just as http has enabled other languages to be used in web applications. This roadmap also suggests plausible HCI studies, including making good UI ubiquitous and reducing design constraints through the integration of web browsers, such as JavaScript, XMLHttpRequest, or even Flash applets. The OpenVanilla team's prototype is an example of such a development, as shown in figure.

---

[1] According to SIGHAN Second International Chinese Word Segmentation Bakeoff's data and documents: http://sighan.cs.uchicago.edu/bakeoff2005/

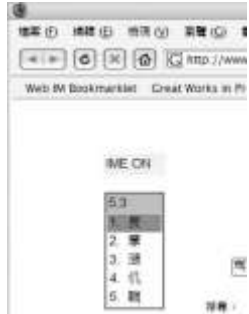

Fig. 5. A Web IM bookmarklet

Since GOING has introduced a two-level candidate window with a matrix in the second level, we shall try to apply both Hick's law [40] and Accot-Zhai's steering law [41]. The former describes the time it takes for a user to make a decision as a function of the possible choices; the latter predicts a user's performance in navigating a hierarchical cascading menu. Menus, or candidate lists, are not the only UI widgets used in IM design. Status windows or other widget forms can also be useful. According to a Microsoft technical report [13], even in the East Asian region, different languages (Japanese, Korean, Chinese) employ different UI schemes, i.e., different combinations of status window or text buffer window (called a composition window by Microsoft). It is funny that even Traditional Chinese and Simplified Chinese IMs employ different schemes. To explain why different languages use different schemes, one may point to differences in language and culture, but this subject requires further investigation. We should also consider the traits common to East Asian languages.

## Conclusion

In this paper, we have covered three major aspects of IM design and implementation, namely, software engineering, human-computer interaction and natural language processing. Various design concepts, such as Lee's NGASR [42], Chang's Open Machine Translation community [43], and the OpenVanilla platform, could help researchers and engineers evaluate their work for real world applications. Ultimately, such research will create more efficient text entry methods for the users.